\documentclass[letterpaper, 10 pt, journal, twoside]{IEEEtran}
\usepackage{cite}

\ifCLASSINFOpdf
  \usepackage[pdftex]{graphicx}
\else
\fi
\usepackage{amsmath}
\usepackage{bm}
\usepackage{algorithm}
\usepackage{algorithmic}

\usepackage{array}
\usepackage{url}

\usepackage{enumitem}
\usepackage{multirow}
\usepackage{multicol}
\usepackage[table]{xcolor}
\DeclareMathOperator{\unit}{unit}

\newif\ifrevision \revisionfalse
\ifrevision
  \newcommand{\rev}[1]{\textcolor{blue}{#1}}
  \newcommand{\revon}{\color{blue}}
\else
  \newcommand{\rev}[1]{#1}
  \newcommand{\revon}{}
\fi

\begin{document}
\bstctlcite{IEEEexample:BSTcontrol}
%
\title{RORA: Realistic Object Reconstruction \\ with Articulation}
\markboth{Under Review for IEEE Robotics and Automation Letters}{}
\author{Hyesung~Lee$^{1}$, Youngseon~Lee$^{2}$, Kyu~Tae~Lee$^{3}$, Dongjun~Lee$^{2}$, and Yongseok~Lee$^{3,*}$%
\thanks{$^{1}$H. Lee is with ETH Z\"{u}rich, Z\"{u}rich, Switzerland.}%
\thanks{$^{2}$Y. Lee and D. Lee are with Seoul National University, Republic of Korea.}%
\thanks{$^{3}$K. T. Lee and Y. Lee are with DGIST, Republic of Korea.}%
\thanks{$^{*}$Corresponding author: Yongseok Lee.}}%
%



\maketitle

\begin{abstract}
Replicating real-world environments into simulation by realistic visual representation such as NeRF and 3D Gaussian Splatting (3DGS) has been an effective strategy to reduce the sim-to-real gap in robot learning. However, bringing articulated objects into simulation is still a challenging task. Existing methods either require capturing the object in multiple joint configurations or rely on learned priors that generalize poorly to multi-joint or out-of-distribution objects. \rev{We propose RORA, a learning-free articulation pipeline that reconstructs simulation-ready assets with articulation from a video scan of a static object, through a suggestion-based human-in-the-loop process.} RORA reconstructs an object with hybrid representation combining mesh and Gaussian for both interactive articulated structure and photorealistic rendering. RORA consists of six main stages: RORA (1) generates initial mesh and Gaussian representation from a static video, (2) refines mesh, (3) decomposes mesh into convex fragments, (4) segments object by grouping the fragments, (5) articulates the object given suggestion, (6) exports the simulatable asset. An Automatic Joint Suggestion Algorithm helps the articulation suggesting candidate joint axes from the local geometry at segmented part boundaries. \rev{On evaluation of 16 objects in PartNet-Mobility dataset, RORA recovered 32 of 33 joints, including chained linkages, with low angular and positional errors. A user study showed that the suggestion algorithm could reduce the articulation cost nearly by half.} We also deploy reconstructed real-world assets in simulation for dexterous manipulation. \footnote{Interactive articulation results and evaluation dataset are available at our project page: \url{https://anonymous.4open.science/w/project-rora-FD49/}}

\end{abstract}


%
\IEEEpeerreviewmaketitle

\begin{figure}[!h]
    \centering
    \includegraphics[width=\columnwidth]{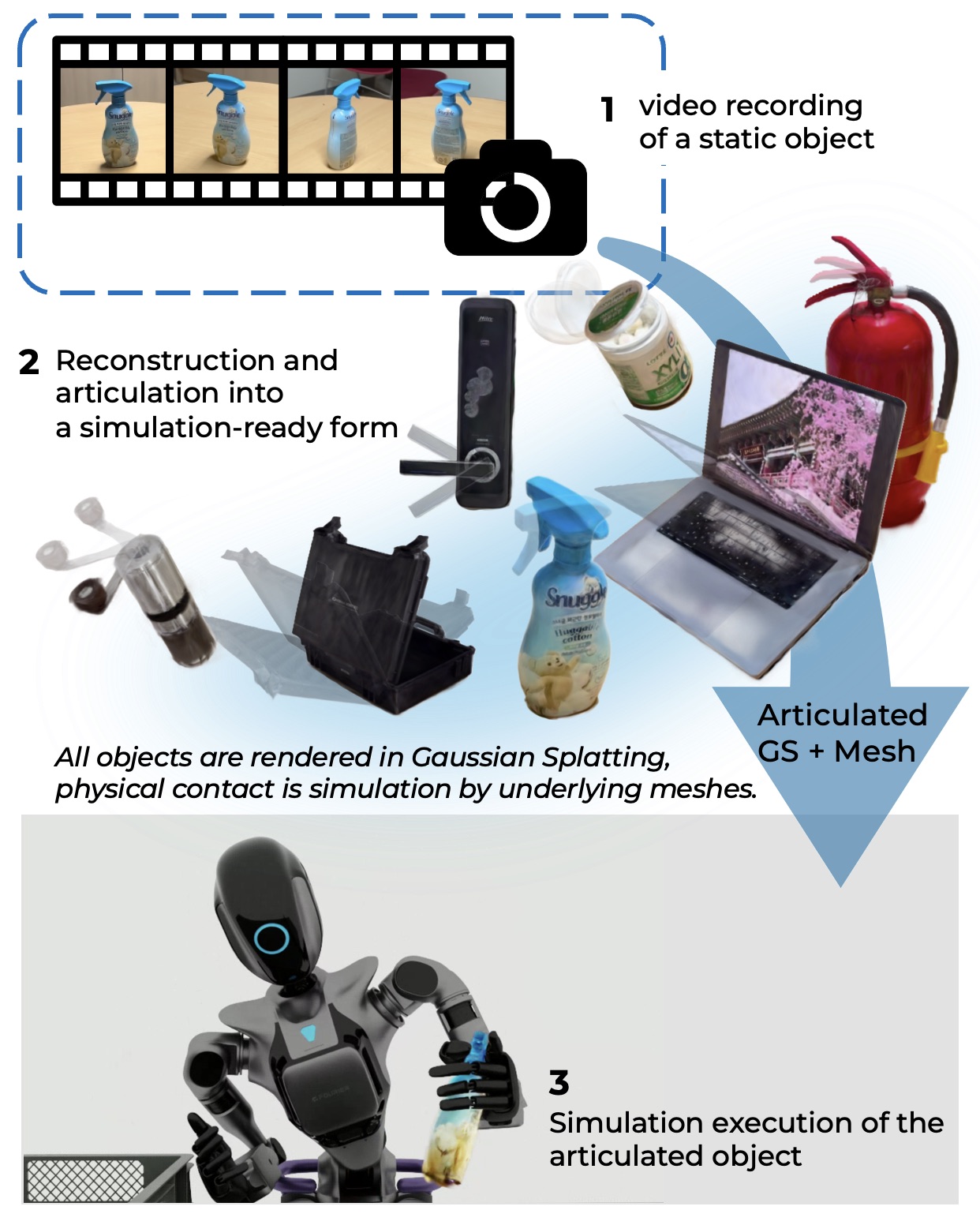}
    \caption{Conceptual overview of RORA framework. Our pipeline reconstructs real objects into simulation-ready, articulated assets. It exports hybrid representation coupling Gaussian Splatting (GS) and meshes. Those articulated assets can easily be imported into simulation environment for robot manipulation tasks.}
    \label{fig:intro_figure}
\end{figure}


\begin{figure*}[!t]
    \centering
    \includegraphics[width=\textwidth]{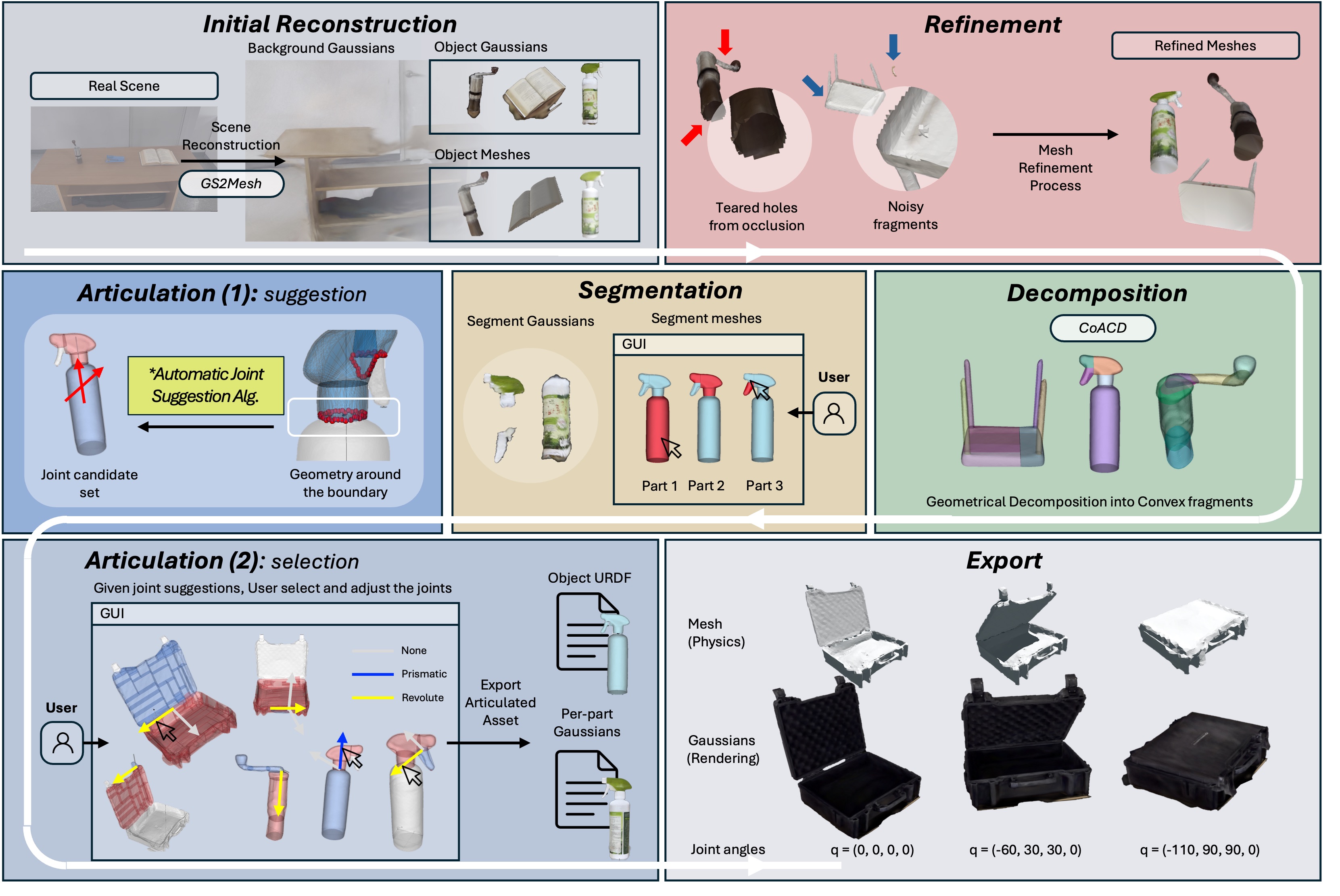}
    \caption{Overall pipeline of our reconstruction framework. The method consists of six stages. \rev{The user acts only at the two stages marked with the user icon: grouping the convex fragments into parts (segmentation) and selecting and fine-tuning one suggested joint per part pair (articulation 2: selection).}}
    \label{fig:pipeline}
\end{figure*}

\section{Introduction}
%
%
%
%

In data-driven robot learning such as humanoid manipulation, replicating a real-world environment in simulation is an effective strategy to reduce the sim-to-real gap, enabling visually realistic and time-efficient data collection~\cite{torne2024rialto, han2026re3sim}. Recent studies suggest that rendering a scene using Gaussian Splatting (GS) can enhance the sim-to-real success rate and improve data efficiency~\cite{qureshi2025splatsim, wu2025rl, li2025robogsim}. Although many real-world objects have internal kinematics, conventional real-to-sim tools cannot infer them, so these objects are represented only as rigid bodies or must be articulated manually one by one.

Many recent studies have addressed this problem. Some studies recover joint kinematics by observing part motion across two or more joint states by optimizing the joints to fit the observations~\cite{jiayi2023paris, kim2025screwsplat}. These methods struggle with complex objects: as the number of joints increases, the number of possible configurations grows exponentially. Learning-based approaches infer articulation either from static object observation or sparse observations by using Vision-Language Models (VLMs)~\cite{le2025articulate}, pretrained diffusion models~\cite{chen2025freeart3d}, or transformer-based networks~\cite{li2026particulate, goyal2025geopard}. However, these models are trained on a limited set of objects and often depend on asset pools. As a result, they do not generalize well to out-of-dataset categories or unseen real-world objects.

We propose \textit{Realistic Object Reconstruction with Articulation (RORA)}, a real-to-sim pipeline that takes a scanning video of a static object as input and exports articulated assets with a hybrid representation of GS and meshes (Fig.~\ref{fig:intro_figure}). RORA uses a suggestion-based human-in-the-loop (HIL) process, since human intuition is important for segmenting and articulating parts with ambiguous cues such as similar color, texture, or geometry~\cite{lee2026human, park2026human}. RORA decomposes the object into fragments and computes candidate joint axes from local geometry information. Then, the user only groups the fragments into parts and selects joint candidates for each part pair, fine-tuning them for higher precision. 

On PartNet-Mobility and on real objects, we show that RORA articulates objects precisely while preserving their appearance, including multi-joint objects and chained linkages. Our contributions are:
{\revon
\begin{enumerate}
\item \textbf{Single static scan:} RORA recovers the kinematic structure from one video of a static object, without moving the joints or capturing several configurations.
\item \textbf{Learning-free joint suggestion:} The \textit{Automatic Joint Suggestion Algorithm} estimates candidate joints from the local geometry at part boundaries. The user accepts or refines them.
\item \textbf{Hybrid simulation-ready assets:} RORA exports a URDF whose links carry a mesh and bound Gaussians, which load directly into Isaac Sim and Unreal Engine.
\end{enumerate}
}


\begin{figure*}[!t]
    \centering
    \includegraphics[width=\textwidth]{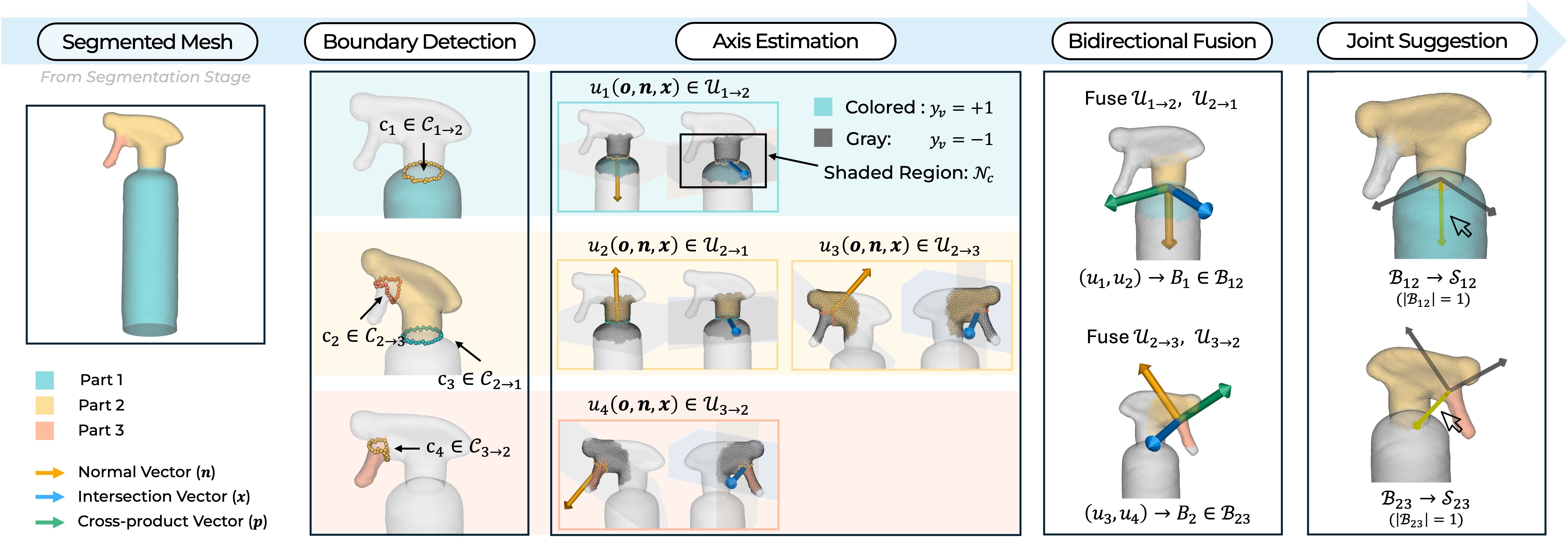}
    \caption{Example of the Automatic Joint Suggestion Algorithm. For each segmented part $i$, boundary cycles $c$ are extracted and added to $\mathcal{C}_{i\to j}$ according to the neighboring part $j$. A unidirectional connection $u\in\mathcal{U}_{i\to j}$ is then computed for each $c\in\mathcal{C}_{i\to j}$, and the connections of the same part pair in $\mathcal{U}_{i\to j}$ and $\mathcal{U}_{j\to i}$ are fused into bidirectional connections $B\in\mathcal{B}_{ij}$. In this example, each part pair meets at a single interface, no pair-level candidate is added, and the axes of the connections in $\mathcal{B}_{ij}$ form the candidate set $\mathcal{S}_{ij}$.}
    \label{fig:articulation_stage}
\end{figure*}

\section{Related Works}

\subsection{Real-to-Sim Reconstruction}
3D Gaussian Splatting (3DGS)~\cite{kerbl2023gaussian} represents real-world scenes with high visual quality, and recent works make it interactive by extracting object meshes~\cite{wolf2024gsmesh} or binding deformable geometry~\cite{jiang2024vrgs}. Several robot learning studies have shown that GS rendering reduces the visual gap between simulation and the real world~\cite{han2026re3sim, qureshi2025splatsim}, but they do not reconstruct articulated objects. \rev{RialTo~\cite{torne2024rialto} supports articulation of mesh objects through a GUI in which the user manually cuts the mesh and places joints, but this requires substantial manual effort and does not give a per-part GS representation. RORA simplifies this procedure with pre-decomposed fragments and joint candidates, and builds movable GS at the same time.}

\subsection{Articulation from Multi-State Observations}
One family of methods articulates objects from images captured in multiple configurations. PARIS~\cite{jiayi2023paris} and DTA~\cite{weng2024neural} estimate a single or multiple joints from multi-view images in two joint states. \rev{ScrewSplat~\cite{kim2025screwsplat} and ArticulatedGS~\cite{guo2025articulatedgs} represent the object with extended 3DGS and jointly optimize part segments and motion. FreeArt3D~\cite{chen2025freeart3d} uses a pretrained 3D diffusion model as a shape prior, and LARM~\cite{yuan2025larm} and ART~\cite{li2026art} recover geometry and joints from sparsely posed images with pretrained transformers. All of these methods require the object to be moved and captured again in each configuration, and some cannot represent kinematic chains; for example, ScrewSplat and ArticulatedGS assume that every movable part is attached to a single static base~\cite{kim2025screwsplat, guo2025articulatedgs}.}

\subsection{VLM and Learning-Based Articulation}
Learning-based methods can infer joints from a single static observation. URDFormer~\cite{chen2024urdformer} constructs a URDF from a single image with a pretrained Vision Transformer, and Real2Code~\cite{mandi2025real2code} reconstructs parts from a few RGB images and lets a language model locate the joints. \rev{Articulate-Anything~\cite{le2025articulate} takes text, images, or a video of the moving joints and runs a VLM actor-critic loop that retrieves part meshes from an asset library and places the joints. Articulate AnyMesh~\cite{qiu2025articulate} prompts a VLM to segment an input mesh into parts and place joints between them. Both rely on semantic priors, so joints are misplaced when a part has no clear semantic label or the retrieved mesh differs from the real object. GEOPARD~\cite{goyal2025geopard} and Particulate~\cite{li2026particulate} take an object mesh and predict joints with a transformer trained on articulated asset datasets. GEOPARD requires the part segmentation as input, while Particulate predicts the kinematic trees from a solid mesh.} These learned models often fail in segmentation and joint placement on real-world objects that their training data does not cover. RORA also works from a single static scan, but it uses no pretrained model or dataset: it infers joint candidates directly from the boundary geometry between parts, and the user confirms the joint type and fine-tunes the placement.


\section{Methods}
RORA has six stages (Fig.~\ref{fig:pipeline}): (1) initial reconstruction, (2) refinement, (3) decomposition, (4) segmentation, (5) articulation, and (6) export. \rev{The user acts at two points only: grouping fragments into parts (Section~\ref{sec:segmentation}) and selecting and fine-tuning one joint candidate per part pair (Section~\ref{sec:selection}).} The other stages are computed automatically. 

\subsection{Initial Reconstruction and Refinement}
\label{sec:initial}
\label{sec:refinement}

We obtain the initial Gaussians and mesh from the RGB video with GS2Mesh~\cite{wolf2024gsmesh}. GS2Mesh uses SAM2~\cite{ravi2024sam2} for per-frame object masks but does not filter the Gaussians themselves, so we keep a Gaussian as part of the object if its center projects inside the mask in more than a set fraction of the frames. The raw mesh has some artifacts: a large hole where the table occluded the object, small tears, and floating fragments. To clean these, we run the refinement stage. We first place the rim of the bottom hole (the largest boundary loop) on the plane $z=0$ and make the mesh watertight in eight steps, where $D_R$ is the bounding-box diagonal: (1) remove fragments smaller than $0.2D_R$; (2) rebuild the surface by screened Poisson reconstruction~\cite{kazhdan2013screened}; (3) remove vertices below $z=0$; (4) decimate to 15k faces; (5) remesh isotropically to an edge length of $0.01D_R$; (6) flatten and cap the bottom; (7) fill the remaining holes; (8) restore vertex colors and normals from the raw mesh.

\subsection{Decomposition and Segmentation}
\label{sec:decomposition}
\label{sec:segmentation}

The mesh is split into non-overlapping convex fragments by approximate convex decomposition (CoACD)~\cite{wei2022approximate}. This stage relies on the assumption that movable parts appear geometrically extruded in the scanned configuration.

Automatic part segmentation often fails when parts share color or texture~\cite{lee2026human}, so the user decides the grouping: the user assigns a part number $i\in\{1,\dots,P\}$ to each fragment by clicking it, with part 1 as the static body. The fragment labels are transferred to the vertices of the refined mesh, giving each vertex $v$ a label $\ell(v)$. A vertex inside a fragment takes the fragment's number, and a vertex outside every fragment takes the label of the nearest labeled vertex along the mesh edges. Each Gaussian is assigned to a part by majority vote among its three nearest mesh vertices, so the mesh and the Gaussians share one part segmentation.

\begin{figure}[!t]
    \centering
    \includegraphics[width=\columnwidth]{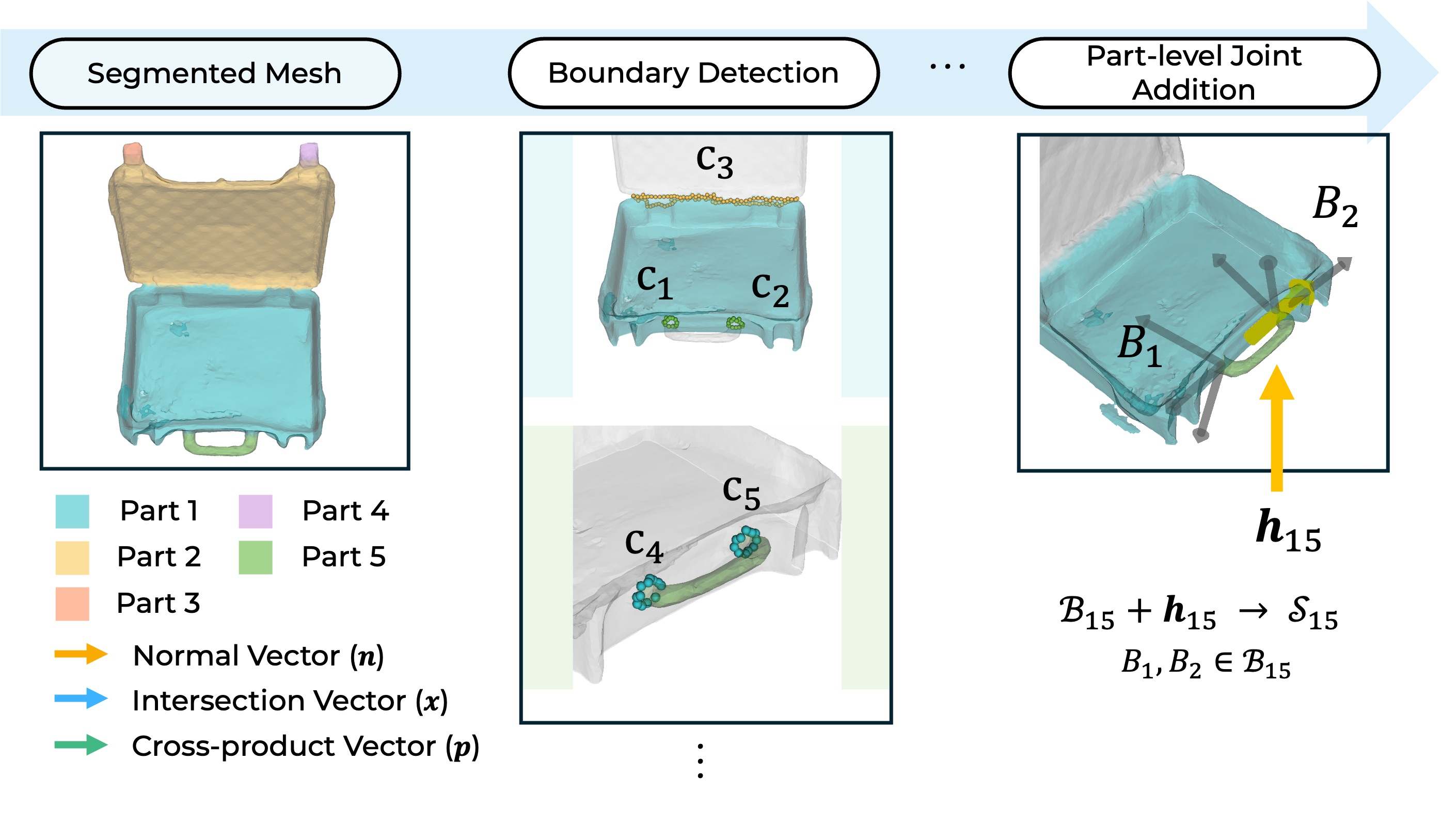}
    \caption{Pair-level candidate between Part 1 and Part 5, which meet at two interfaces. From the two bidirectional connections $B_1, B_2\in\mathcal{B}_{15}$, the pair-level candidate $\bm{h}_{15}$ is placed along the line through their origins and added to the candidate set $\mathcal{S}_{15}$.}
    \label{fig:articulation_additional}
\end{figure}

\begin{algorithm}[t]
    \caption{\rev{Automatic Joint Suggestion Algorithm}}
    \label{alg:suggestion}
    \renewcommand{\algorithmiccomment}[1]{\hfill\textit{// #1}}
    \begin{algorithmic}[1]
    \small
    \FOR{each side $(s,t)\in\{(i,j),(j,i)\}$}
        \STATE $\mathcal{C}_{s\to t}\leftarrow$ boundary cycles of $s$ connected to $t$ \COMMENT{III-C1}
        \FOR{each cycle $c\in\mathcal{C}_{s\to t}$}
            \STATE $\bm{o}\leftarrow$ centroid of $c$, $\bm{n}\leftarrow$ SVM normal \eqref{eq:svm} \COMMENT{III-C2}
            \STATE $\bm{x}\leftarrow\unit(\bm{n}_1\times\bm{n}_2)$ of the better two-plane EM fit
            \STATE add $u=(\bm{o},\bm{n},\bm{x})$ to $\mathcal{U}_{s\to t}$
        \ENDFOR
    \ENDFOR
    \STATE $\mathcal{B}_{ij}\leftarrow$ pairs matched by $c_{ab}$ \eqref{eq:cost}, fused by \eqref{eq:fuse} \COMMENT{III-C3}
    \STATE $\mathcal{S}_{ij}\leftarrow\{(\bm{o},\bm{d})\mid B\in\mathcal{B}_{ij},\ \bm{d}\in\{\bm{n},\bm{x},\bm{p}\}\}$
    \IF[III-C4]{$|\mathcal{B}_{ij}|\ge 2$}
        \STATE add the pair-level candidate $\bm{h}_{ij}$ to $\mathcal{S}_{ij}$
    \ENDIF
    \RETURN $\mathcal{S}_{ij}$ \COMMENT{suggested to the user}
    \end{algorithmic}
\end{algorithm}

\subsection{Articulation Stage}

Once segmentation is complete, the \textit{Automatic Joint Suggestion Algorithm} proposes candidate joint axes for every pair of adjacent parts (Fig.~\ref{fig:articulation_stage} and Alg.~\ref{alg:suggestion}). \rev{The algorithm assumes the following: a movable part is attached to its parent through a contact interface, and the shape of that interface constrains the joint axis. A revolute or prismatic axis either runs along the normal of a flat interface or lies along the crease line where two parts meet. The algorithm therefore fits the local geometry of each interface with one or two separating planes and suggests their normal, their intersection line, and the direction perpendicular to both as candidates.} It consists of four steps: boundary detection, axis estimation, bidirectional fusion, and pair-level candidate addition. $G=(V,E,F)$ denotes the refined mesh with vertices $V$, edges $E$, and faces $F$; $\bm{r}_v$ is the position of vertex $v$, and $D_A$ is the diagonal of the refined mesh bounding box.

\subsubsection{\textbf{Boundary Detection}}

A \textit{boundary edge} of a set of faces is an edge that belongs to only one face of the set. For each part $i$, let $F_i\subseteq F$ be the faces whose three vertices all have label $i$. Since the faces are consistently oriented, a boundary edge of $F_i$ is a directed edge $(u,v)$ of a face in $F_i$ whose reverse $(v,u)$ is not in $F_i$. Starting from an unvisited boundary edge, we repeatedly move to the boundary edge leaving the current end vertex until we return to the start. Each cycle becomes a \textit{boundary cycle} $c$ of part $i$.  When the labeling is noisy, one physical interface can be split into several cycles. We therefore sort the cycles by length in descending order and keep a cycle only if more than 50\% of its edges are not contained in the cycles kept before. A cycle $c$ of part $i$ is connected to part $j$, and added to the set $\mathcal{C}_{i\to j}$, if at least 25\% of its vertices share an edge with a vertex of part $j$.

\subsubsection{\textbf{Axis Estimation}}

For each boundary cycle $c \in \mathcal{C}_{i\to j}$, the system computes a \textit{unidirectional connection} $u=(\bm{o}, \bm{n}, \bm{x})$, consisting of an origin $\bm{o}$, the centroid of the cycle vertices, and a normal vector $\bm{n}$ and an intersection vector $\bm{x}$ estimated from the local geometry. The neighborhood of $c$ is the set of vertices within $K=10$ edges of the cycle, $\mathcal{N}_c=\{v\in V \mid \min_{u\in c} \mathrm{dist}_G(u,v)\le K\}$, where $\mathrm{dist}_G$ is the edge distance on $G$. Since the remeshed triangles are nearly uniform in size, $\mathcal{N}_c$ collects the vertices nearby along the surface. In $\mathcal{N}_c$, vertices of part $i$ are labeled $y_v=+1$ and those of any other part $y_v=-1$, and separating planes between the two classes are fitted.

{\revon
\textit{Single-plane estimation.} A flat interface is described by one plane. We fit a linear support vector machine (SVM)~\cite{cortes1995support} to the labeled neighborhood $\mathcal{N}_c$:
\begin{equation}
  \min_{\bm{w},\,b}\; \tfrac{1}{2}\lVert\bm{w}\rVert^2 + C \sum_{v\in\mathcal{N}_c} \max\bigl(0,\; 1 - y_v(\bm{w}^\top\bm{r}_v + b)\bigr)^2
  \label{eq:svm}
\end{equation}
with $C=1$, and take the unit normal of the separating plane, $\bm{n}=\bm{w}^*/\lVert\bm{w}^*\rVert$, where $(\bm{w}^*, b^*)$ is the minimizer.
}

{\revon
\textit{Multi-plane estimation.} A bent interface has a crease, which one plane cannot represent. We fit two planes $f_k(\bm{r})=\bm{w}_k^\top\bm{r}+b_k$, $k\in\{1,2\}$, to $\mathcal{N}_c$ and describe the side of part $i$ as either convex or concave: in the convex model, part $i$ occupies the intersection of the two positive half-spaces; in the concave model, it occupies their union. Each model is fitted with the hard-assignment expectation-maximization (EM) algorithm~\cite{dempster1977maximum}. In the E-step of the convex model, every positive vertex is assigned to both planes and each negative vertex to the plane with the most negative value, $k^*=\arg\min_k f_k(\bm{r}_v)$; the concave model does the opposite, assigning each positive vertex to $k^*=\arg\max_k f_k(\bm{r}_v)$ and every negative vertex to both planes. In the M-step, each plane is refitted with~\eqref{eq:svm} on its assigned vertices, with each class weight $C$ set to the inverse of its number of assigned vertices so that both classes contribute equally. The alternation stops when the assignment no longer changes or after 20 iterations, and we keep the model with the higher classification accuracy on $\mathcal{N}_c$. With the unit normals $\bm{n}_1, \bm{n}_2$ of its two planes, the intersection vector is $\bm{x} = \unit(\bm{n}_1\times\bm{n}_2)$. If the planes are nearly parallel ($\lVert\bm{n}_1\times\bm{n}_2\rVert<10^{-6}$), the boundary is treated as planar and no intersection vector is computed.
}

\subsubsection{\textbf{Bidirectional Fusion}}

Because boundaries are detected per part (ideally twice), the interface between parts $i$ and $j$ is seen from both sides, and each side can have a different number of unidirectional connections. {\revon The system matches the two unidirectional connection sets $\mathcal{U}_{i\to j}$ and $\mathcal{U}_{j\to i}$ one to one with the Hungarian matching algorithm~\cite{kuhn1955hungarian}, minimizing the sum of the pair costs
\begin{equation}
  c_{ab} = w_p\,\lVert\bm{o}_a-\bm{o}_b\rVert/D_A + w_a\bigl(1-\lvert\bm{n}_a^\top\bm{n}_b\rvert\bigr)
  \label{eq:cost}
\end{equation}
over $a\in\mathcal{U}_{i\to j}$ and $b\in\mathcal{U}_{j\to i}$, with $w_p=1.0$ and $w_a=0.5$. Each matched pair $(a,b)$ is fused into a bidirectional connection $B=(\bm{o}, \bm{n}, \bm{x}, \bm{p})$ after aligning the signs:
\begin{equation}
\begin{aligned}
  \bm{o} &= \tfrac{1}{2}(\bm{o}_a+\bm{o}_b), & \bm{n} &= \unit(\bm{n}_a + s_n\bm{n}_b),\\
  \bm{x} &= \unit(\bm{x}_a + s_x\bm{x}_b), & \bm{p} &= \unit(\bm{n}\times\bm{x}),
\end{aligned}
\label{eq:fuse}
\end{equation}

where $s_n=\operatorname{sgn}(\bm{n}_a^\top\bm{n}_b)$ and $s_x=\operatorname{sgn}(\bm{x}_a^\top\bm{x}_b)$. The vector $\bm{p}$ is perpendicular to both $\bm{n}$ and $\bm{x}$ and covers axes that lie in the interface but across the crease. The three vectors, all through $\bm{o}$, are the candidate axes of the connection. The sign of a candidate only decides the positive direction of the joint coordinate. When the two sides have different numbers of connections, unmatched connections are discarded, except when only one side has a connection. That single connection forms a bidirectional connection without averaging.
}

\subsubsection{\textbf{Pair-Level Candidate Addition}}
When a part pair meets at several interfaces, the joint axis is constrained by all of them together, so we add a pair-level candidate. Let $\mathcal{B}_{ij}$ be the bidirectional connections between parts $i$ and $j$. If $|\mathcal{B}_{ij}|=2$, a rotation that keeps both interfaces in place has its axis through both origins, so the candidate $\bm{h}_{ij}$ is placed at their midpoint along the line through them (e.g., the toolbox handle in Fig.~\ref{fig:articulation_additional}). If $|\mathcal{B}_{ij}|\ge 3$, $\bm{h}_{ij}$ is placed at the mean of the origins along the mean of the sign-aligned normals. Writing a candidate as an axis line $(\bm{o},\bm{d})$, the candidate set shown to the user is $\mathcal{S}_{ij}=\{(\bm{o},\bm{d}) \mid B\in\mathcal{B}_{ij},\ \bm{d}\in\{\bm{n},\bm{x},\bm{p}\}\}\cup\{\bm{h}_{ij}\}$. The additional vector ${\bm{h}_{ij}}$ is included into $\mathcal{S}_{ij}$ only if $|\mathcal{B}_{ij}|\ge 2$.

\subsubsection{\textbf{Joint Selection}}
\label{sec:selection}
This is the only user step in the articulation stage. The GUI goes through the part pairs $(i,j)$ whose candidate set $\mathcal{S}_{ij}$ is not empty and shows the two parts with their candidate axes. The user selects a candidate, sets its type to revolute or prismatic, and fine-tunes its origin and direction if needed. A pair with no selected candidate gets no joint.

{\revon
The system then forms the selected joints into a kinematic tree automatically. Part 1 becomes the root, and a breadth-first search from the root visits neighbors in increasing part number. Each joint is directed from parent to child in the direction in which the search first reaches it. Since URDF describes only trees, the selection is dropped if the joints form a closed loop, or if a part pair has more than one joint unless these joints together form a planar or spherical joint. A part that the search does not reach is attached to the base with a fixed joint, and the tree is written as a URDF file.
}

\subsection{{Export}}
The export stage writes the URDF with one mesh and one Gaussian file per link. Each link mesh, opened at its interfaces after segmentation, is closed again by hole filling so that it can be imported directly into a simulator, and the simulator can render Gaussian file with the rigid transform bound to its link to keep the photorealistic rendering. \rev{By default, the range of a revolute joint is $[-\pi, \pi]$ and that of a prismatic joint is $[-D_A, D_A]$, the effort and velocity limit is 10 and 1 respectively. The mass of each link is computed from its mesh volume with a constant density.}


\begin{table}[t]
\centering
\caption{Kinematic Metrics Evaluation}
\label{tab:benchmark}
\setlength{\tabcolsep}{2pt}
\resizebox{\columnwidth}{!}{%
\begin{tabular}{l|ccc|ccc|ccc}
\hline
 & \multicolumn{3}{c|}{\textbf{Cat.~I}} & \multicolumn{3}{c|}{\textbf{Cat.~II}} & \multicolumn{3}{c}{\textbf{Cat.~III}} \\
\textbf{Method} & Found$\uparrow$ & Ang.$\downarrow$ & Pos$\downarrow$ & Found$\uparrow$ & Ang.$\downarrow$ & Pos$\downarrow$ & Found$\uparrow$ & Ang.$\downarrow$ & Pos$\downarrow$ \\
\hline
AA~\cite{le2025articulate}                     & 5/6 & 3.07 & 1.67 & 14/17 & 16.43 & 9.01 & 9/10 & 40.04 & 26.81 \\
\rev{AM}~\cite{qiu2025articulate}                    & 4/6 & 27.20 & 23.75 & 13/17 & 44.12 & 15.06 & 5/10 & 21.07 & 9.01 \\
\rev{Particulate}$^\dagger$~\cite{li2026particulate} & 5/6 & 3.95 & 0.84 & \textbf{17/17} & 0.97 & 1.40 & 9/10 & \textbf{0.37} & \textbf{1.71} \\
ScrewSplat$^\ddagger$~\cite{kim2025screwsplat} & 3/6 & \textbf{0.12} & \textbf{0.10} & 7/17 & \textbf{0.29} & 6.07 & 4/10 & 0.53 & 40.25 \\
\hline
\rowcolor{gray!15}
\rev{RORA-Sugg}                                         & \textbf{6/6} & 1.63 & 0.86 & 16/17 & 8.57 & 2.22 & \textbf{10/10} & 2.21 & 3.49 \\
\rowcolor{gray!15}
RORA-Full                                           & \textbf{6/6} & 0.74 & 0.59 & 16/17 & 1.64 & \textbf{0.69} & \textbf{10/10} & 0.89 & 2.04 \\
\hline
\end{tabular}}
\\[2pt]
\parbox{\columnwidth}{\scriptsize The unit of Found joints, angular error, positional error is count, degree, \% respectively. \rev{$^\dagger$Particulate is privileged since its training dataset includes the partnet-mobility dataset. $^\ddagger$ Chains (Cat.~III) are outside of ScrewSplat's single-base model assumption.}}
\end{table}

\begin{table}[t]
\centering
\caption{Visual Metric Evaluation}
\label{tab:visual_grouped}
\setlength{\tabcolsep}{2.5pt}
\resizebox{\columnwidth}{!}{%
\begin{tabular}{ll|ccc|ccc}
\hline
 & & \multicolumn{3}{c|}{\textbf{SSIM}} & \multicolumn{3}{c}{\textbf{LPIPS}} \\
\textbf{Objects} & \textbf{Method} & Rest$\uparrow$ & $\Delta$ mean$\uparrow$ & $\Delta$ std$\downarrow$ & Rest$\downarrow$ & $\Delta$ mean$\downarrow$ & $\Delta$ std$\downarrow$ \\
\hline
 & AA~\cite{le2025articulate}           & \textbf{0.929} & $-0.036$ & 0.036 & \textbf{0.075} & $+0.016$ & 0.025 \\
 & ScrewSplat~\cite{kim2025screwsplat}  & 0.867 & $-0.002$ & 0.047 & 0.159 & $+0.007$ & 0.046 \\
\rowcolor{gray!15}
\cellcolor{white}\multirow{-3}{*}{\shortstack{Cat.~I\\(6 obj.)}} & RORA                                 & 0.914 & $\mathbf{+0.012}$ & \textbf{0.024} & 0.089 & $\mathbf{-0.003}$ & \textbf{0.018} \\
\hline
 & AA~\cite{le2025articulate}           & 0.763 & $-0.054$ & 0.086 & 0.236 & $+0.057$ & 0.092 \\
 & ScrewSplat~\cite{kim2025screwsplat}  & 0.773 & $-0.007$ & \textbf{0.023} & 0.242 & $\mathbf{+0.008}$ & \textbf{0.042} \\
\rowcolor{gray!15}
\cellcolor{white}\multirow{-3}{*}{\shortstack{Cat.~II\\(6 obj.)}} & RORA                                 & \textbf{0.862} & $\mathbf{-0.001}$ & 0.040 & \textbf{0.129} & $+0.010$ & 0.043 \\
\hline
 & AA~\cite{le2025articulate}           & 0.585 & $\mathbf{+0.004}$ & \textbf{0.046} & 0.389 & $+0.033$ & 0.054 \\
 & ScrewSplat~\cite{kim2025screwsplat}$^\ddagger$ & 0.654 & $-0.014$ & 0.055 & 0.372 & $\mathbf{+0.029}$ & \textbf{0.045} \\
\rowcolor{gray!15}
\cellcolor{white}\multirow{-3}{*}{\shortstack{Cat.~III\\(4 obj.)}} & RORA                                 & \textbf{0.883} & $-0.075$ & 0.136 & \textbf{0.138} & $\mathbf{+0.029}$ & 0.075 \\
\hline
\end{tabular}}
\par\smallskip
{\footnotesize Rest: zero state. $\Delta$: articulated minus zero-state over random configurations (mean, std).
\raggedright \rev{$^\ddagger$Category~III is outside ScrewSplat's single-base model.} \par}
\end{table}

\section{Experiments}
\label{sec:experiments}

We evaluated RORA in three parts: a comparison with baselines on a synthetic benchmark (Section~\ref{sec:benchmark}), ablations of the joint suggestion algorithm and of the suggestions themselves (Section~\ref{sec:ablation}), and reconstruction of real objects and their deployment in simulators (Section~\ref{sec:real}).

\subsection{Benchmark Comparison}
\label{sec:benchmark}

\noindent\textbf{Evaluation dataset.}
We used 16 objects from PartNet-Mobility~\cite{xiang2020sapien} with 33 movable joints, grouped by kinematic structure. \textit{Category~I} has a single joint (6 objects), \textit{Category~II} has multiple joints on one fixed base (6 objects), and in \textit{Category~III} at least one joint has a moving link as its parent (4 objects). The object IDs are available on the project page.

\noindent\textbf{Baselines and inputs.} {\revon
Each method received the input it is designed for.
RORA received 3D Gaussians trained on 48 views of one static configuration and a voxelized mesh (voxel pitch 
$0.01\,m$) of the merged GT parts.
Articulate-Anything (AA)~\cite{le2025articulate} accepts several input modalities but performs best on video, so we gave it the GT part meshes and a 56-frame video in which all joints moved together, with up to 10 actor-critic iterations per GT joint.
Articulate AnyMesh (AM)~\cite{qiu2025articulate} and Particulate~\cite{li2026particulate} segment and articulate a static mesh, and both received the same voxelized mesh as RORA.
AA and AM used GPT-6-luna as the VLM.
All 16 objects are in Particulate's training split, so its scores are not on held-out data ($\dagger$). ScrewSplat~\cite{kim2025screwsplat} received 48 views in each of 5 joint states (240 images).
Category~III is outside its single-base assumption ($\ddagger$).
}

\begin{figure*}[!t]
    \centering
    \includegraphics[width=\textwidth]{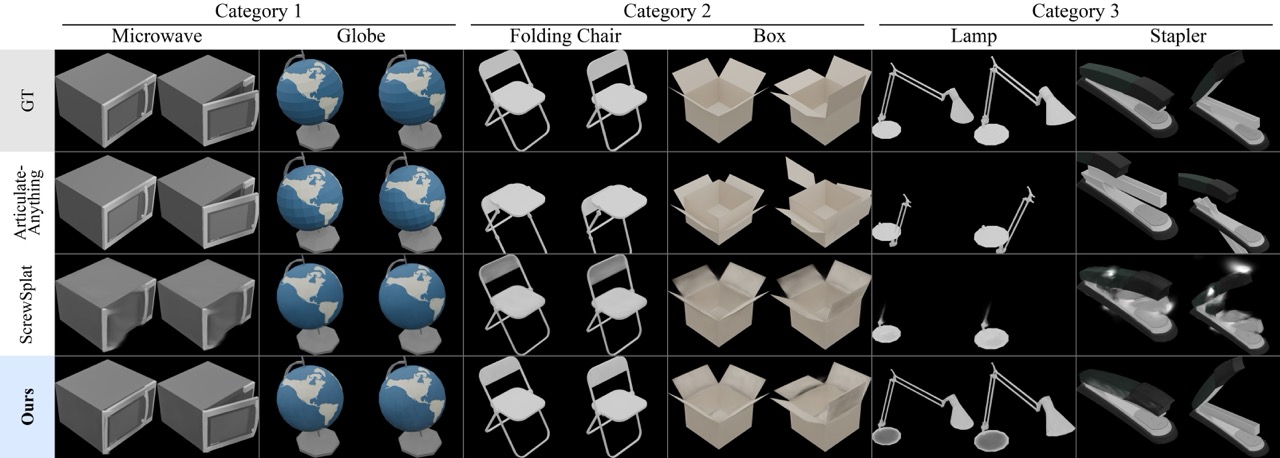}
    \caption{Rendering comparison on the PartNet-Mobility benchmark, two examples per category. For each method, the left image is the zero configuration and the right image is the articulated state used for evaluation against the ground truth (GT).}
    \label{fig:overall_comparison}
\end{figure*}

{\revon
\noindent\textbf{RORA variants.}
To separate the algorithm from the human correction, we report two variants. \textit{RORA-Sugg} scores the candidate set $\mathcal{S}_{ij}$ before any user input on joints: for each GT joint between parts $(i,j)$, we take the candidate in $\mathcal{S}_{ij}$ with the smallest angular error, which measures the best suggestion the user can choose. \textit{RORA-Full} is the exported asset after one of the authors selected the type and adjusted the joints, taking less than a minute per object. 21 of the 32 found joints were adjusted after selection.
}

\subsubsection{\textbf{Kinematic metrics}}
{\revon
Each GT joint, with origin $\bm{o}$ and unit axis $\bm{a}$, was paired with the semantically corresponding predicted joint $(\hat{\bm{o}},\hat{\bm{a}})$, if one existed. We measured three metrics:
\begin{itemize}[leftmargin=*, nosep]
    \item \textit{Found joints}: the number of GT joints that have a predicted joint of the same type between the same pair of links. Extra predictions are not counted.
    \item \textit{Angular error}: $e_\text{ang}=\arccos\lvert\bm{a}^\top\hat{\bm{a}}\rvert\in[0^\circ,90^\circ]$.
    \item \textit{Positional error}: the relative distance from $\hat{\bm{o}}$ to the GT axis line, $e_\text{pos}=\frac{100}{D}\lVert(\hat{\bm{o}}-\bm{o})\times\bm{a}\rVert$ [\%], computed only for revolute joints. $D$ is the box diagonal of the zero-state GT object.
\end{itemize}
}

\begin{table}[t]
\centering
\caption{\rev{Ablation of the automatic joint suggestion algorithm}}
\label{tab:ablation}
\setlength{\tabcolsep}{3.5pt}
\small
\begin{tabular}{l|cccc}
\hline
Variant & Found$\uparrow$ & Ang ($^\circ$)$\downarrow$ & Pos (\%)$\downarrow$ & Miss $\downarrow$ \\
\hline
Single-plane only & \textbf{32/33} & 44.20 & \textbf{2.44} & 17 \\
Multi-plane only  & \textbf{32/33} & 55.48 & 2.45 & 22 \\
w/o cross         & \textbf{32/33} & 22.39 & 2.53 & 10 \\
w/o fusion        & 30/33 & \textbf{3.40} & 3.41 & \textbf{4} \\
\hline
\rowcolor{gray!15}
All (RORA-Sugg)     & \textbf{32/33} & 5.28 & \textbf{2.44} & \textbf{4} \\
\hline
\end{tabular}
\\[2pt]
\parbox{\columnwidth}{\scriptsize Miss: no joint found or $e_{ang}>15^\circ$}
\end{table}

\providecommand{\std}[1]{{\scriptsize$\pm$#1}}

\begin{table}[t]
\centering
\caption{\rev{Ablation of Joint editing with and without suggestions}}
\label{tab:userstudy}
\setlength{\tabcolsep}{3pt}
\small
\begin{tabular}{l|cc|cc}
\hline
 & \multicolumn{2}{c|}{\textbf{Cost}} & \multicolumn{2}{c}{\textbf{Accuracy}} \\
Condition & Time (s)$\downarrow$ & Inputs$\downarrow$ & Ang ($^\circ$)$\downarrow$ & Pos (\%)$\downarrow$ \\
\hline
Manual     & 103.6 \std{56.7} & 59.1 \std{28.0} & \textbf{1.62} \std{1.05} & 1.60 \std{0.85} \\
\rowcolor{gray!15}
Suggested  & \textbf{56.0} \std{28.6} & \textbf{32.2} \std{18.6} & 2.61 \std{1.62} & \textbf{1.35} \std{0.74} \\
\hline
\end{tabular}
\\[2pt]
\parbox{\columnwidth}{\scriptsize Values are mean $\pm$ std across the 15 participants, each first averaged over their two objects in that condition.}
\end{table}

{\revon
Angular and positional errors are averaged only over the found joints, so the three columns have to be read together. Since the candidates of RORA-Sugg have no type yet, a GT joint between parts $(i,j)$ counts as found for RORA-Sugg if $\mathcal{S}_{ij}$ contains a corresponding candidate. 
}

Table~\ref{tab:benchmark} shows the kinematic accuracy of joints by methods. Note that both error metrics are averaged over the found joints only, a method that finding fewer joints can show lower errors. ScrewSplat placed joints accurately, with low angular errors in Cat.~I, II, but found at most half of the joints in every category. The errors of AA grew from Cat.~I to Cat.~II, III as the kinematic structure became more complex. AM had low found rates and large errors in all categories, which makes it unreliable for articulation. Particulate found most joints with high accuracy, but since its training data includes the test objects, we use it only as a privileged reference. Both RORA methods found 32 of the 33 joints and kept the positional error near the minimum in all three categories (at most 3.49\%). With the suggestions alone, RORA-Sugg kept the angular error below $2.5^\circ$ except in Cat.~II ($8.57^\circ$), where a few joints had large errors. Human fine-tuning easily reduced these errors, and RORA-Full kept the angular error below $2^\circ$ in all categories.

\begin{figure*}[!t]
    \centering
    \includegraphics[width=\textwidth]{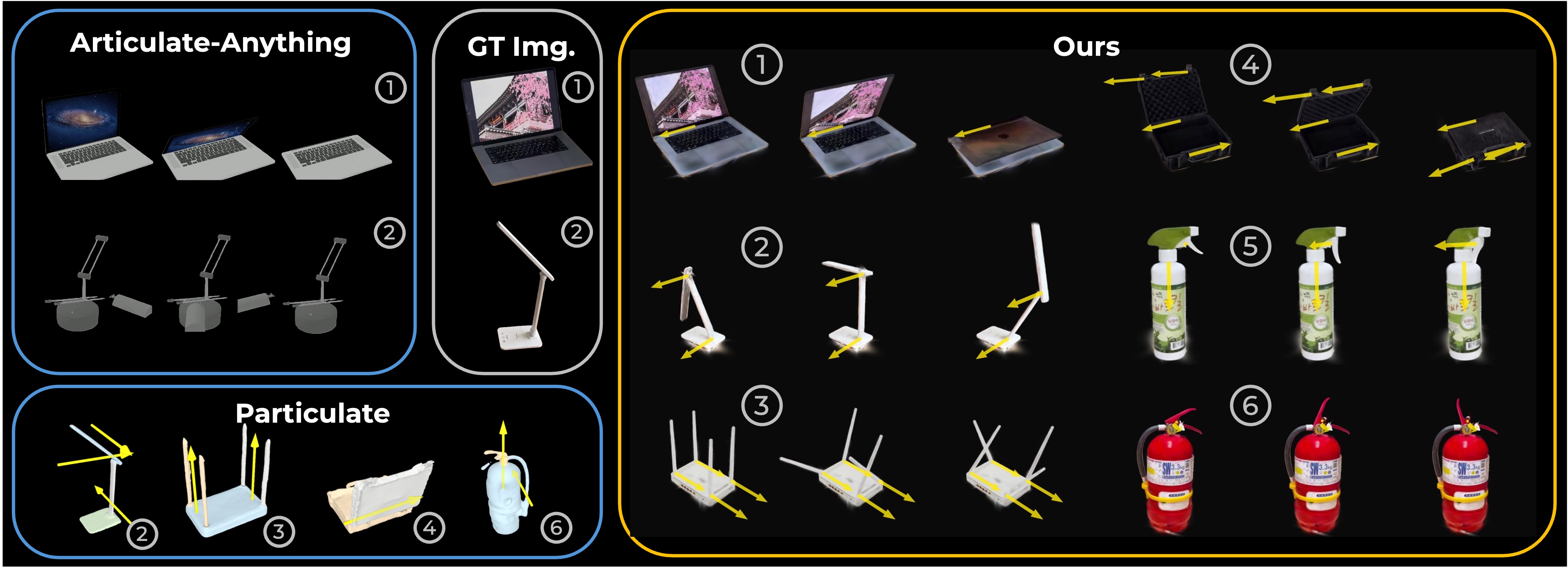}
    \caption{Qualitative comparison of Articulate-Anything, Particulate, and RORA on real-world objects. Photographs of objects 1 and 2 are shown for reference. The circled numbers index the objects. Yellow arrows indicate revolute joint axes.}
    \label{fig:real_objects}
\end{figure*}

\begin{figure}[!t]
    \centering
    \includegraphics[width=\columnwidth]{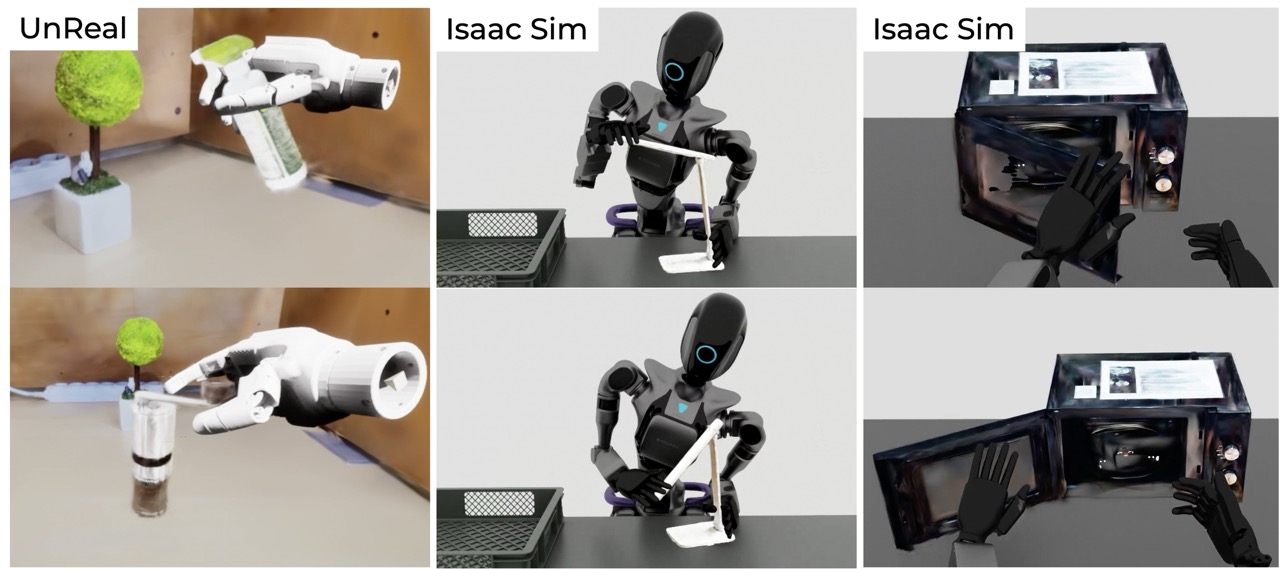}
    \caption{Dexterous manipulation in simulation: a robot hand triggers the spray in Unreal Engine (left) and a humanoid robot bends a desk lamp in Isaac Sim (middle), the same robot opens the microwave in Isaac Sim (right).}
    \label{fig:simulation}
\end{figure}

    \subsubsection{\textbf{Visual metrics}}
We compared the renders of AA, ScrewSplat, and RORA-Full by SSIM and LPIPS (AlexNet). AM and Particulate were excluded because they articulate a given mesh and do not reconstruct appearance from images or video. {\revon Renders are cropped to the smallest square around the GT object, resized to $512 \times 512$, and scored only on the union of the object pixels of the GT and all methods, a mask shared by all methods. We report the score in the zero configuration of the GT object and the mean change ($\Delta$) to three random configurations.} The sample images are in Fig.~\ref{fig:overall_comparison}.

\rev{In Table~\ref{tab:visual_grouped}, the zero-state score reflects reconstruction quality, and $\Delta$ reflects articulation accuracy, i.e., how much the appearance changes against GT as the joints move.} RORA has almost the highest zero-state scores in all categories, since it represents most links and places them at the correct position. AA, although given the GT part meshes, misplaces parts and does well only on Category~I. As the objects become more complex in Category~II, III, the zero-state scores of AA drop sharply, and its $\Delta$ also becomes worse. $\Delta$SSIM in Category ~III even appears positive because the zero-state render is already far from GT. The scores of ScrewSplat also drop sharply from Category~I to II. In contrast, RORA shows little degradation under articulation even in the complex categories, with high zero-state score, and small $\Delta$ in Categories~I and II. In Category~III, its $\Delta$SSIM drops more ($-0.075$), but its articulated scores remain high. Overall, RORA has consistentently high zero-state score with small $\Delta$ mean, which shows its visual reliability in articulation.

\subsection{Ablations}
\label{sec:ablation}

\subsubsection{\textbf{Components of the joint suggestion algorithm}}
We ran the algorithm on the same segmentation with one component changed and scored the candidates in the same way as RORA-Sugg (Table~\ref{tab:ablation}). \rev{The bidirectional connection $B=(\bm{o}, \bm{n}, \bm{x}, \bm{p})$ of \textit{Single-plane only} condition keeps the normal $\bm{n}$ of each connection, \textit{multi-plane only} keeps the intersection vector $\bm{x}$, \textit{w/o cross} omits the cross-product vector $\bm{p}$, and \textit{w/o fusion} omits the bidirectional fusion and uses only the connections seen from the parent side.}

With single-plane only, multi-plane only, and w/o cross, the algorithm missed 17, 22, and 10 over 33 joints (\textit{miss} means no joint found or $e_{ang}>15^\circ$). \rev{ This is because each vector covers a different type of joint: $\bm{n}$ captures prismatic joints and rotations about a flat contact face, $\bm{x}$ captures hinge joints, and $\bm{p}$ captures joints between very thin parts, where the other two fail. Without bidirectional fusion, the found joints decreased to 30/33, because connections are detected asymmetrically.} The lower angular error of this variant is a survivorship effect, as the two joints with large errors are omitted from the average, and its positional error increased from 2.44\% to 3.41\%, since fusion corrects the origin by averaging the two sides.

\subsubsection{\textbf{Effect of suggestions on users}}
{\revon To test whether the suggestions reduce effort at comparable accuracy, we compared the time, the number of inputs (clicks and key presses), and the articulation accuracy of manual joint placement without suggestions (\textit{Manual}) and with suggestions (\textit{Suggested}, as in RORA-Full). Fifteen participants with no prior experience in articulation took part after a 20-minute tutorial. Individual participant articulated two objects with each method in the same GUI. In the Manual condition, all other conditions are identical except: a triad of $xyz$ axes is created at the origin, and the user sets the parent and child parts. The objects were segmented in advance so that only articulation was measured. The order of the conditions was balanced, no participant saw the same object again, and the 15 objects were equally distributed to all participants. Before each articulation, participants watched a video of the original URDF in motion, and each object was rotated about the $z$-axis by a random angle between $15^\circ$ and $75^\circ$ so that the triad at the origin could not match a joint axis by chance.
}

{\revon With suggestions, the time per object and the number of inputs both dropped by almost half (Table~\ref{tab:userstudy}). Accuracy stayed low-level in both conditions, since participants adjusted each joint by until it satisfied their own judgment. The positional error stayed similar, while the angular error was slightly higher: when a suggested axis deviated by only a few degrees, the difference was hard to see in the GUI, and participants often accepted the candidate without adjusting it. The difference in the mean angular error ($1.62^\circ$ vs.\ $2.61^\circ$) was not significant of either condition.}

\subsection{Real-World Asset Generation and Simulated Robot Manipulation}
\label{sec:real}

As shown in Fig.~\ref{fig:real_objects}, our pipeline reconstructs real-world objects photorealistically, preserving fine visual elements such as text labels and surface textures, and the segmented 3DGS and URDF meshes move according to the joint constraints of the real objects. Articulate-Anything fails on these objects because the retrieved meshes do not match the out-of-dataset objects, Particulate failed to infer correct joint given our refined mesh as input.

To test the assets for data collection in simulation, we deployed them on two platforms (Fig.~\ref{fig:simulation}). In the GPU-accelerated simulator~\cite{ji2025gpu} built on Unreal Engine 5.3, operators teleoperated an Inspire Hand through a visual-inertial hand tracking interface~\cite{lee2021vist}. In NVIDIA Isaac Sim, bi-manual hands of a Fourier GR-1 were controlled in real time with Meta Quest controllers. In both, operators could interact with the reconstructed objects and collect demonstration data.

\section{Conclusion}

In this paper, we presented a whole real-to-sim pipeline with learning-free articulation algorithm, which reconstructs real-world objects with articulation from a single scan of a static object into a hybrid representation of 3D Gaussian Splats and meshes. Users get a simulation-ready asset with minimal input of grouping fragments into parts and selecting suggested joints. The experiments showed our method found the most joints (32/33) with errors comparable to the best baselines, and the highest zero-state visual scores in Categories II and III. The assets were used for dexterous manipulation in Unreal Engine and NVIDIA Isaac Sim. We expect that this framework reduces the manual work of asset creation for robot simulation.

Our framework has some limitations. Since the part segmentation and joint suggestion rely on geometric features, they fail to distinguish movable parts that are not extruded or are extremely small relative to the whole object. Sharing the fundamental limitation of Gaussian Splatting, the initial reconstruction fails on reflective or transparent surfaces and \rev{cannot generate surfaces occluded during the scan.} \rev{In future work, we plan to infer more physical properties such as mass, stiffness, and friction at the joints, to use the reconstructed assets for training humanoid robots with real-to-sim transfer,} and to host an asset sharing platform for the robotics community.

\ifCLASSOPTIONcaptionsoff
  \newpage
\fi



%

\bibliographystyle{IEEEtran}
\bibliography{references}

%








\end{document}